\documentclass{article}
\usepackage{arxiv}

\usepackage[utf8]{inputenc} % allow utf-8 input
\usepackage[T1]{fontenc}    % use 8-bit T1 fonts
\usepackage{url}            % simple URL typesetting
\usepackage{booktabs}       % professional-quality tables
\usepackage{amsfonts}       % blackboard math symbols
\usepackage{nicefrac}       % compact symbols for 1/2, etc.
\usepackage{microtype}      % microtypography
\usepackage{lipsum}
\usepackage{authblk}
\usepackage[misc]{ifsym}
\usepackage{graphicx}
\usepackage{multicol}
\usepackage{multirow}
\usepackage{amsmath}
\usepackage[colorlinks,linkcolor=blue]{hyperref}

\title{Med3D: Transfer Learning for 3D Medical Image Analysis}
\author[1]{Sihong Chen\thanks{whalechen@tencent.com}}
\author[1]{Kai Ma}
\author[1]{Yefeng Zheng}
\affil[1]{Tencent YouTu X-Lab, Shenzhen, China}

\begin{document}
\maketitle

\begin{abstract}
The performance on deep learning is significantly affected by volume of training data. Models pre-trained from massive dataset such as ImageNet become a powerful weapon for speeding up training convergence and improving accuracy. 
Similarly, models based on large dataset are important for the development of deep learning in 3D medical images. However, it is extremely challenging to build a sufficiently large dataset due to difficulty of data acquisition and annotation in 3D medical imaging.
We aggregate the dataset from several medical challenges to build 3DSeg-8 dataset with diverse modalities, target organs, and pathologies. To extract general medical three-dimension (3D) features, we design a heterogeneous 3D network called Med3D to co-train multi-domain 3DSeg-8 so as to make a series of pre-trained models.
We transfer Med3D pre-trained models to lung segmentation in LIDC dataset, pulmonary nodule classification in LIDC dataset and liver segmentation on LiTS challenge. 
Experiments show that the Med3D can accelerate the training convergence speed of target 3D medical tasks 2 times compared with model pre-trained on Kinetics dataset, and 10 times compared with training from scratch as well as improve accuracy ranging from 3\% to 20\%. Transferring our Med3D model on state-the-of-art DenseASPP segmentation network, in case of single model, we achieve 94.6\% Dice coefficient which approaches the result of top-ranged algorithms on the LiTS challenge.
\end{abstract}

%%%%%%%%%%%%%%%%%%%%%%training_efficiency%%%%%%%%%%%%%%%%%%%%
\begin{figure}
\begin{center}
\includegraphics[width=0.9\textwidth]{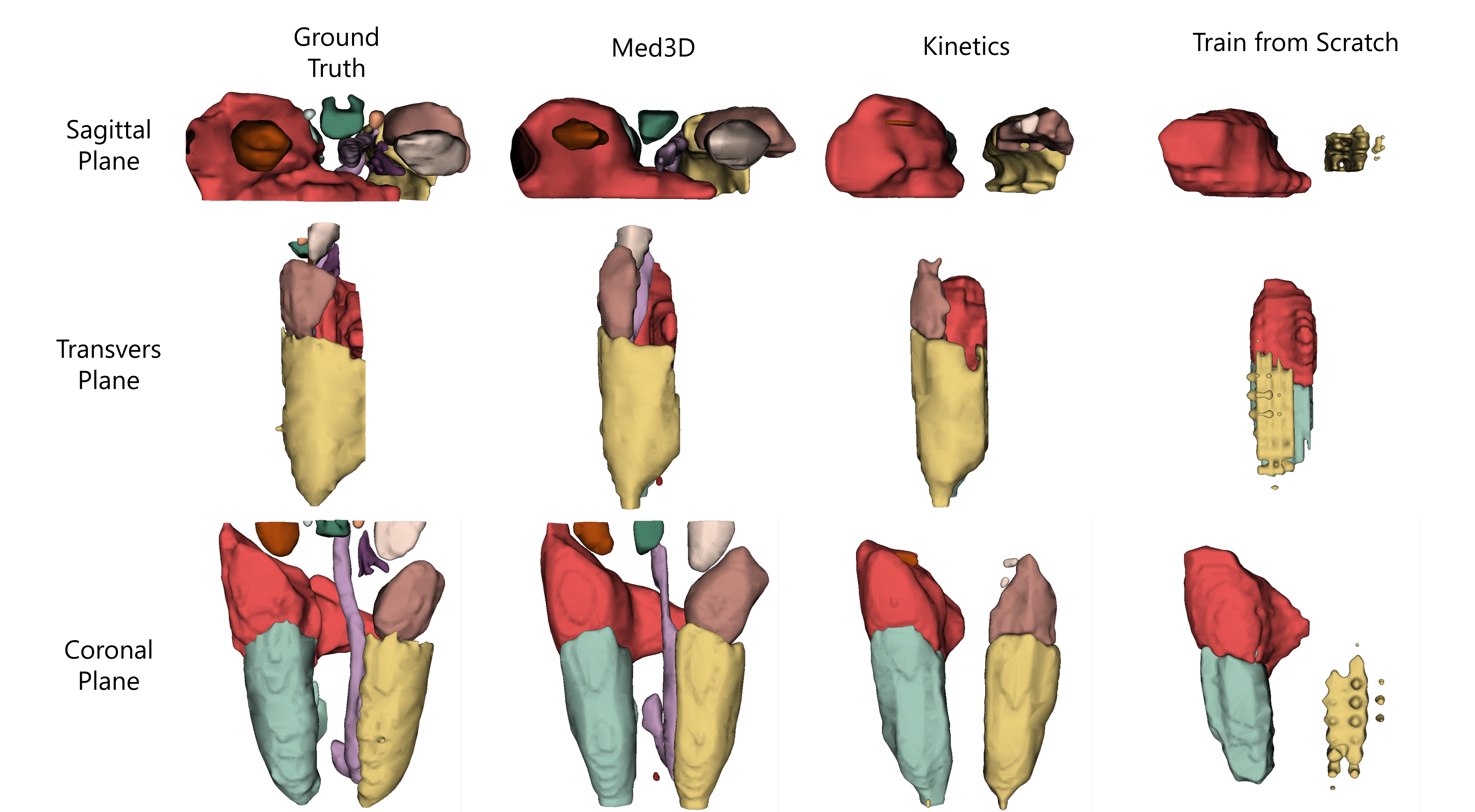}
\end{center}
   \caption{Visualization of the segmentation results of our approach vs. the comparison ones after the same training epochs.}
\label{fig1}
\end{figure}
%%%%%%%%%%%%%%%%%%%%%%training_efficiency%%%%%%%%%%%%%%%%%%%%
\section{Introduction}
Data driven approaches, e.g. deep convolutional neural networks (DCNN), recently have achieved state-of-the-art performance on different vision tasks, such as image classification, semantic segmentation, object detection, etc. It is known that one of the fundamental facts contributing to such success is the massive amount of training data with detailed annotations. Suppose taking natural image domain for example, the ImageNet dataset~\cite{russakovsky2015imagenet} consists more than 14 million images with over 20 thousand categories and MS COCO dataset~\cite{lin2014microsoft} collects more than a million images with rich instance segmentation annotations.

In the medical imaging domain, however, it is extremely challenging to build a sufficiently large 3D dataset due to the intrusive nature of some medical imaging modalities (e.g. CT), the prolonged imaging duration as well as the laborious annotation in 3D. As a consequence, there has not been a large-scale 3D medical dataset that is publicly available to train baseline 3D DCNNs. To avoid the inferior performance caused by training networks from scratch using a small set of data, some studies~\cite{han2017automatic,yu2018recurrent} converted 3D volume data to 2D and leveraged the pre-trained 2D models from ImageNet~\cite{russakovsky2015imagenet}. Although this solution gains better performance than training from scratch, there still a big gap due to the abandoned 3D spatial information. Some other methods tried to utilize the 3D spatial information by initializing with networks trained from the Kinetics dataset~\cite{kay2017kinetics}. However, the 3D information captured by the temporal video data and the medical volume is so different that there exists a strong bias for learning a 3D medical image network transferred from natural scene videos. 

In this work, we try to solve the aforementioned problems from two aspects, building a large 3D medical dataset, and training a baseline network with such data that can be transferred to solve other medical problems. In the first step, we collect many small-scale datasets from different medical domains with various imaging modalities, target organs and pathological manifestations, to build a relatively large 3D medical image dataset. %We also employ a general data normalization method for unifying multi-domain dataset to have consistent spatial and intensity distributions. 
In the second step, we establish an encode-decode segmentation network called Med3D that is heterogeneously trained with the aforementioned large-scale 3D medical dataset. A multi-branch decoder is also proposed to tackle the incomplete annotation problem. 
We also transfer the extracted universal encoder of Med3D to lung segmentation, nodule classification and LiTS challenge~\cite{lits}, compared to 3D models based on the Kinetics video dataset and models trained from scratch. Experiments have demonstrated that the efficiency for training convergence and accuracy based on our proposed pre-trained models (see Fig.~\ref{fig1}).

The contributions of our work are summarized as follows: (1) We propose a heterogeneous Med3D network aiming for 3D multi-domain medical data, which can extract general 3D features even in the case of large differences of data domain distribution. (2) We transfer backbone of the Med3D model to three new 3D medical image tasks. And, we confirmed the effectiveness and efficiency of Med3D with extensive experiments. (3) To facilitate the community to reproduce our experimental results and apply Med3D to other applications, we will release our Med3D pre-trained models and relative source code\footnote[1]{Pre-trained models and relative source code has been made publicly available at \href{https://github.com/Tencent/MedicalNet}{https://github.com/Tencent/MedicalNet}.}.

%%%%%%%%%%%%%%%%%%%%%%introduction%%%%%%%%%%%%%%%%%%%%

%%%%%%%%%%%%%%%%%%%%%%related work%%%%%%%%%%%%%%%%%%%%
\section{Related Work}
Sun et al.~\cite{sun2017revisiting} have shown that the greater the amount of data, the better the performance of deep learning networks. In natural images, people have been building lots of large-amount datasets for decades, such as ImageNet~\cite{russakovsky2015imagenet}, PASCAL VOC~\cite{everingham2010pascal}, MS COCO~\cite{lin2014microsoft}, etc., which include abundant annotations on millions of images. Pre-trained models based on these large-scale datasets can extract useful general features which are widely used in classification, detection, and segmentation tasks. Studies~\cite{ravishankar2016understanding,tajbakhsh2016convolutional,erhan2009difficulty,yanai2015food,he2018rethinking} have repeatedly proved that pre-trained models can accelerate the training convergence speed and increase the accuracy of the target model.

Similarly, large-scale dataset and corresponding pre-trained models are important in medical imaging applications. Over the past decade, over 100 challenges~\cite{grand} have been organized within the area of biomedical image analysis~\cite{zhuang2019evaluation, raudaschl2017evaluation, armato2011lung, jimenez2016cloud, menze2015multimodal}. Most of these challenges are 3D semantic segmentation tasks, and the data magnitude is limited which are generally in the tens or hundreds, much fewer than natural images. 
Therefore, each dataset is too small to stably pre-train a 3D model for transfer learning. In this work, we aggregate many small 3D datasets to build a large 3DSeg-8 dataset for pre-training.
%3D data is widely used in the medical field.
In view of the limited dataset and the absence of 3D medical image pre-trained model, Han et al.~\cite{han2017automatic} sliced 3D data from three axes for the purpose of using a model pre-trained from natural image; Yu et al.~\cite{yu2018recurrent} tried to analyze the third dimension with a temporal approach in recurrent neural network~\cite{zhao20183d} and transfer models based on natural scene video~\cite{hara2018can} to 3D network. However, such methods still cannot fully leverage the structural information of the third dimension. Most of the studies on 3D medical imaging such as ~\cite{yang2016automatic, milletari2016v, cciccek20163d} prefer to train a small 3D convolution neural network from scratch.

Pan and Yang~\cite{pan2010survey} have indicated that the more similar the data distribution between source domain and target domain, the better the transferring effect. According to the above observation the above defects, we believe that the pre-trained model based on 3D medical dataset should be superior to natural scene video in 3D medical target tasks. Due to the lack of large-scale 3D medical images, co-trained models based on multi-domain 3D medical image dataset may be a solution. Training network cross different domains simultaneously is a challenging task. Duan et al.~\cite{duan2009domain} proposed a data-dependent regularizer based on support vector machine for video concept detection. Hoffman et al.~\cite{hoffman2012discovering} introduced a mixture-transform model for object classification and Nam et al.~\cite{nam2016learning} presented a domain adaptation network for video tracking. Since the pixel presentation and range of pixel value of medical imaging on different domains are completely different, the above methods for natural images cannot be directly applied onto the medical imaging. 

%-------------------------------------------------------------------------
%%%%%%%%%%%%%%%%%%%%%%related work%%%%%%%%%%%%%%%%%%%%

%-------------------------------------------------------------------------
%%%%%%%%%%%%%%%%%%%%%%framework%%%%%%%%%%%%%%%%%%%%
\begin{figure*}
\begin{center}
\includegraphics[width=0.9\textwidth]{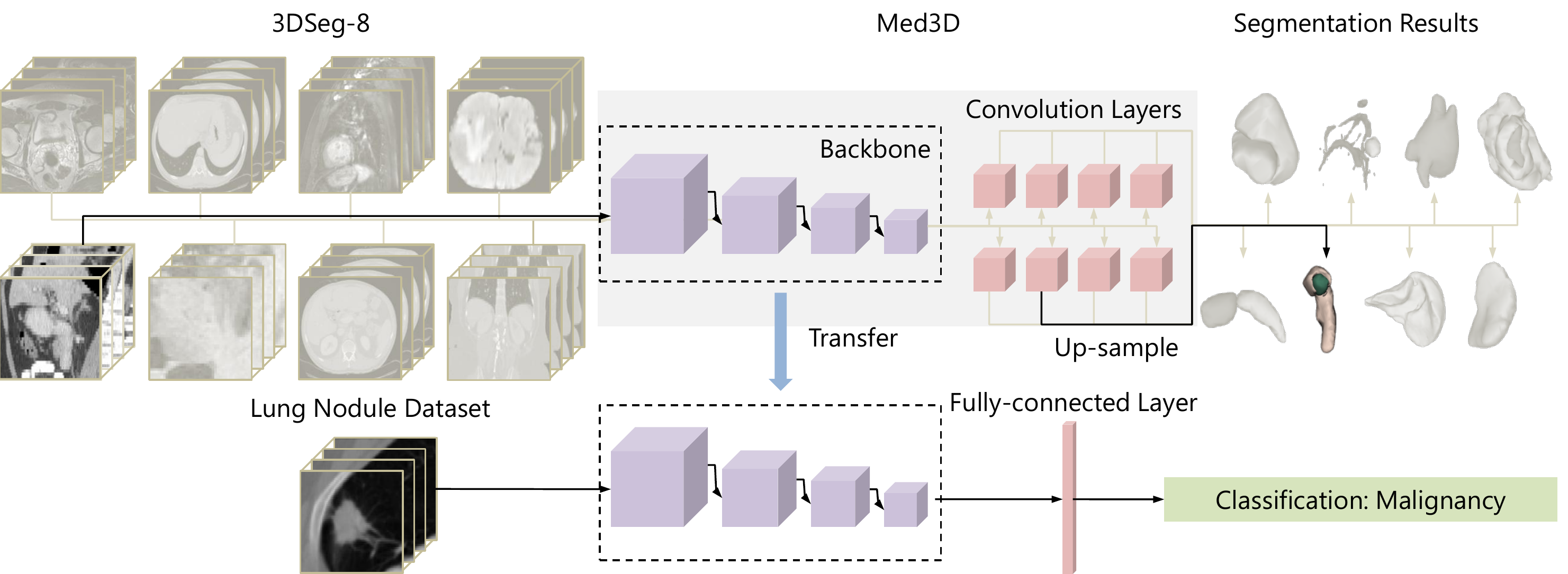}
\end{center}
   \caption{Framework of the proposed method.}
\label{fig2}
\end{figure*}
%%%%%%%%%%%%%%%%%%%%%%framework%%%%%%%%%%%%%%%%%%%%
%-------------------------------------------------------------------------
%%%%%%%%%%%%%%%%%%%%%%method%%%%%%%%%%%%%%%%%%%%
\section{Method}
The motivation of our work is to train a high performance DCNN model with a relatively large 3D medical dataset, that can be used as the backbone pre-trained model to boost other tasks with insufficient training data. To reach this target, we design a processing workflow with three major steps, as shown in Fig.~\ref{fig2}. In the first step, we collect several publicly available 3D segmentation datasets called 3DSeg-8 from different medical imaging modalities, e.g. magnetic resonance imaging (MRI) and computed tomography (CT), with various scan regions, target organs and pathologies. We then normalize all the data with the same spatial and intensity distributions. In the second step, we train a DCNN model, namely Med3D, to learn the features. The network has a shared encoder and eight simple decoder branches for each specific dataset. In the last step, the extracted features from the pre-trained Med3D model are transferred to other medical tasks to boost the network performance. Details of each step are explained in the following sections. 

\subsection{Data Selection and Normalization}
Our data is collected from eight different 3D medical segmentation datasets and we refer the composed dataset as 3DSeg-8 for later convenience. The reason why we select data from segmentation datasets rather than classification ones are mainly two folds: firstly, unlike the natural image analysis that can learn general feature representations from thousands of object categories, the medical imaging analysis operates on a confined body region that has much less object classes. Learning from a small set of classes in a classification task may lead to a poor generalization. Secondly, the classification label turns to be much weaker supervision information for 3D medical data compared to natural image labels, as the label may only correspond to a tissue/organ existing in a small portion of the volume data, which may hamper the learning process of the neural network. %For example, the tumor usually has a size less than 5cm, where the liver scan protocol usually cover the full abdominal region. 
Therefore, we select data from those eight segmentation datasets and believe that learning the tissue/organ differences from the segmentation task could result in better representative features.

The original volume data of 3DSeg-8 is from different modalities (MRI and CT), distinctive scan regions (e.g. brain, heart, prostate, spleen, etc.) as well as multi-centers, which create a large variety of data characteristics, such as the spatial resolutions in 3D and range of pixel intensities. To mitigate the data variation problem that may hamper the network training process, we process the data with a pre-processing module that accomplishes the spatial and intensity distribution normalization.

\textbf{Spatial Normalization}. Medical volumes are common with heterogeneous voxel spacing resulting from different scanners or different acquisition protocols. Such spacing refers to the physical distance between two pixels in an image which is almost completely different from domain to domain. More importantly, such spacing information can not be learned from CNN.
Spatial normalization (i.e., resampling volumes to a fixed resolution) is often employed to reduce the effect of voxel spacing variation. To avoid over interpolation, we interpolate each volume to the median spacing according to the respective domain so as to keep the spatial characters of target in each domain. In domain $j$, the spacings from each axis of x, y and z in the $i$th image can be defined as $sp_{i, j}^{ax}$ where $ax$ belongs to ${x, y, z}$. The median spacings of the $j$th domain data is calculated as:

~\begin{equation*}
    sp_{med, j}^{ax} = f_{med}(sp_{0, n}^{ax}, sp_{1, n}^{ax}, ..., sp_{N_{j}, n}^{ax}),
\end{equation*}
where $f_{med}$ denotes the median operation and $N_{j}$ presents the number of data in $j$th domain. And, the new image size $s_{i, j}^{ax}$ can be calculated from size in original image $s_{ax, i, j}$ as below:

~\begin{equation}
    s_{ax, i, j}^{'} = \frac{s_{ax, i, j}}{sp_{i, j}^{ax}}\times sp_{med, j}^{ax}.
\end{equation}

Since the target organ of each domains a different proportion of the whole image, after spacing normalization, we randomly crop the region with size varies from two times of the target bounding box to the whole volume for ensuring that targets are fully contained in the training data.

\textbf{Intensity Normalization}. 
Under different imaging modalities, the range of pixel values of medical images varies significantly. To eliminate the side-effect of pixel value outliers, especially for the CT modality (eg. metal in CT), we sort all the pixel values in each image and truncate the intensities to the range of 0.5 to 99.5 percentiles. Due to the different intensity range from various domains, we further normalize intensity value $v_{i}$ to $v_{i}^{'}$ using the mean $v_{m}$ and standard deviation $v_{sd}$ in an individual volume as:

~\begin{equation}
    v_{i}^{'} = \frac{v_{i} - v_{m}}{v_{sd}}.
\end{equation}

\subsection{Med3D Network}
The current popular pre-trained networks, such as ResNet ~\cite{he2016deep} and pre-activation ResNet~\cite{he2016identity}, haven been proven to be effective to transfer to other vision tasks, as they offer a universal encoder, covering both low- and high-level features that are also helpful in other domains. Therefore, in the design of Med3D network, we leverage the existing feature extraction architecture of the mature networks and focus on exploring the optimal way to train Med3D with our unique 3DSeg-8 data. As the goal is to learn universal feature representations by training a segmentation network, the common encode-decode segmentation architecture is adopted, where the encoder can be any basic structures. In this work, we adopt the family of ResNet models as the basic structure of the encoder and do minor modifications to allow the network to train with 3D medical data.
One big difference between 3DSeg-8 data and data in natural image segmentation tasks is that the multi-organ segmentation annotations are missing\footnote[2]{There exists 3D medical dataset with multi-organ segmentation annotations with small amount of data such as TCIA (53 samples)~\cite{clark2013cancer} and BTCV~\cite{marsh2013beyond} (47 samples).} where only the particular organ/issue of interest in each dataset is annotated as the foreground class and others are left to be the background. For example, only liver segmentation masks are available in the liver/tumor dataset and the nearby pancreas is marked as the background~\cite{lits}. Similarly, only the pancreas is marked as the foreground in the pancreas/tumor dataset~\cite{MSD}. Such incomplete annotation information could confuse the network and the training process does not converge. 

Since it is technically impossible to detailedly annotate the complete organ atlas in large-scale 3D medical data, we propose a multi-branch decoder to tackle the incomplete annotation problem. As shown in Fig.~\ref{fig2}, we connect the encoder with eight specific decoder branches, where each one of them corresponds to one particular dataset of 3DSeg-8. At the training stage, each decoder branch only processes the feature map extracted from the corresponding dataset and the rest branches do not participate in its optimization process. Moreover, each decoder branch share the same architecture of using one convolution layer to combine features from the encoder. To compute the differences between the network output and the ground truth annotation, we directly up-sample the feature map from the decoder to the original image size. Such simple decoder design allows the network to focus on training a universal encoder. At the test stage, the decoder part is removed and the remained encoder can be transferred to other tasks.

\subsection{Transfer Learning}
Our goal is to establish a general 3D backbone network that can be transferred to other medical tasks to gain better performance than training from scratch. To validate the effectiveness and versatility of the established Med3D network, we conduct three different and thorough experiments in segmentation and classification domains. 

\textbf{Lung Segmentation}. 
Med3D only sees different tissues in the local receptive fields and is never surpervised by lung annotation. In order to verify whether the features generated by the Med3D are universal in the lung segmentation task on large receptive fields (whole human body), we transfer encoder part from Med3D as the feature extraction part and then segmented lung in whole body followed by three groups of 3D decoder layers. The first set of decoder layers is composed of a transposed convolution layer with a kernel size of $(3, 3, 3)$ and a channel number of 256 (which is used to amplify twice the feature map), and the convolutional layer with $(3, 3, 3)$ kernel size and 128 channels. The remaining two groups of decoder layers are similar to the first group excep doubling the number of channels per layer of network progressively. Finally, the convolution layer in the $(1, 1, 1)$ kernel is employed to generate the final output and the number of channels corresponds to the number of categories.

\textbf{Pulmonary Nodule Classification}. To further verify the versatility of Med3D on non-segmentation tasks, as shown in Fig.~\ref{fig2}, we transfer the encoder part of Med3D to the nodule classifier as a feature extractor, and the average pooling operation and fully-connected layer with $(1, 1, 1)$ kernel size is added to classify the results. Different to human organs (e.g., spleen and pancreas), a pulmonary nodule is a micro-structure with a much smaller size. Furthermore, pulmonary nodules are in character with blurred boundaries, large deformation of shape and rich texture information which are quite different compared with normal human organs. Transferring the Med3D model to the pulmonary nodule task can help to examine whether the features produced by Med3D are also versatile for microscopic tissue or not.

%%%%%%%%%%%%%%%%%%%%%%liver framework%%%%%%%%%%%%%%%%%%%%
\begin{figure}
\begin{center}
\includegraphics[width=0.7\textwidth]{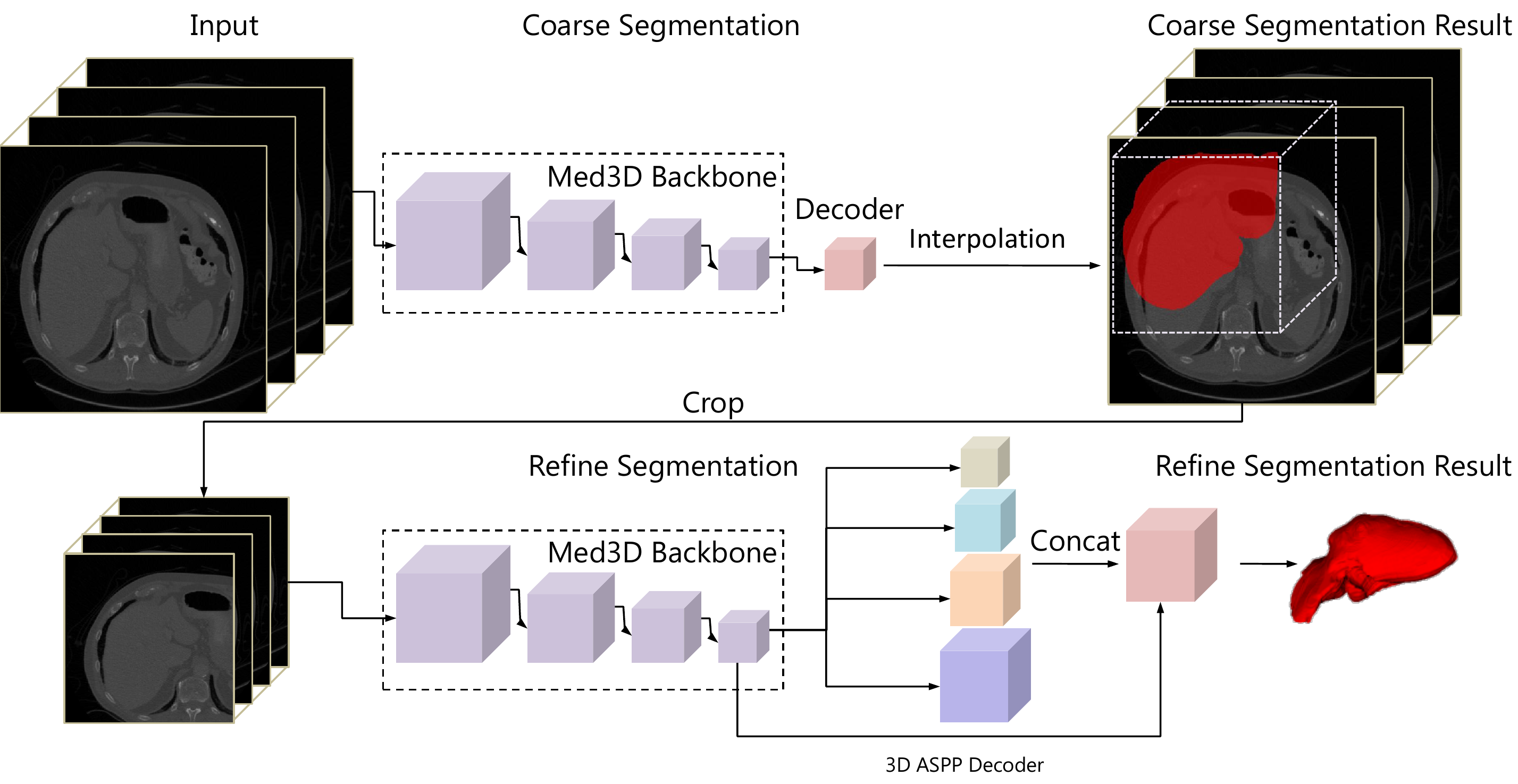}
\end{center}
   \caption{Framework of the liver segmentation.}
\label{fig3}
\end{figure}
%%%%%%%%%%%%%%%%%%%%%%liver framework%%%%%%%%%%%%%%%%%%%%

\textbf{LiTS Challenge}. 
To evaluate the performance of Med3D pre-trained model in practical applications, we apply the Med3D model to the liver segmentation in the LiTS competition~\cite{lits}. The liver is a common site of primary tumor development. Low intensity contrast to liver tumor and adjacent tissues presents a big obstacle to accurately segment the liver. The liver is a single connected area the same as left atrium. Inspired by the method from first place winner of the 2018 3D Atrial Segmentation Competition~\cite{xiong2018fully}, we employ the two-stage segmentation network to segmenting liver.

As illustrated in Fig.~\ref{fig3}, firstly, we roughly segment liver in the whole image to obtain the region of interest (ROI) of the target. In this stage, we transfer the backbone pre-trained from Med3D as the encoder part, followed by a convolution layer with $(1,1,1)$ kernel and the number of channels is two (liver vs. background). After resizing image, we input the image to coarse-segmentation network for extracting features with 32 times downsampling. Then, we upsample the feature map to the original image size by a bi-linear interpolation method.

Secondly, we crop the liver target area according to the result from first stage and sub-segment the target again so as to obtain the final liver segmentation result. This stage is mainly focused on careful liver contour segmentation. In order to obtain more dense scale information in the feature map and obtain a larger receptive field, we embed the backbone pre-trained from Med3D into the state-the-of-art DenseASPP segmentation network. This network connects a set of atrous convolutional layers in a dense way which generates multi-scale features that not only cover a larger scale range, but also cover that scale range densely without significantly increasing the model size. And, we replace all the 2D kernels with the corresponding 3D version. Since there is inevitably bias between ground truth and liver target prediction in the result from the first step, in order to increase robustness of the fine-segmentation model, we perform a random expansion liver target area and then process it with two augmentation methods including rotation and translation.

%%%%%%%%%%%%%%%%%%%%%%method%%%%%%%%%%%%%%%%%%%%

%%%%%%%%%%%%%%%%%%%%%%experiment%%%%%%%%%%%%%%%%%%%%
\section{Experiments}
In this section, we conduct several experiments to explore the performance of the proposed Med3D models. First, we show detailed settings of the experiments and some comparison results with different dataset settings. Next, we transfer the pre-trained Med3D encoder to initialize networks of other medical tasks and compare the results with training-from-scratch networks and pre-trained networks with Kinetics data. Finally, we concatenate the pre-trained Med3D encoder with DenseASPP ~\cite{yang2018denseaspp} network and demonstrate the state-of-the-art performance on the liver segmentation task using a single model. 

%%%%%%%%%%%%%%%%%%%%%%MSD data details%%%%%%%%%%%%%%%%%%%%
\begin{table}
\begin{center}
\caption{Details of 3DSeg-8 where ``Tumours" and ``Tissue" represent segmentation of lesions and components in an organ (``Organ").}
\begin{tabular}{c|c|c|c}
\hline
Dataset & \#Cases & Modality & Target Type \\ \hline\hline
Brain~\cite{menze2015multimodal} & 485 & MRI & Tumour \\ \hline
Hippo~\cite{MSD} & 261 & MRI & Tissue \\ \hline
Prostate~\cite{MSD} & 33 & MRI & Tissue \\ \hline
Liver~\cite{lits} & 131 & CT & Organ \\ \hline
Heart~\cite{tobon2015benchmark} & 100 & MRI & Tissue \\ \hline
Pancreas~\cite{MSD} & 282 & CT & Organ/Tumour \\ \hline
Vessel~\cite{MSD} & 304 & CT & Tissue/Tumour \\ \hline
Spleen~\cite{MSD} & 42 & CT & Organ \\ \hline
\end{tabular}
\label{tab1}
\end{center}
\end{table}
%%%%%%%%%%%%%%%%%%%%%%MSD data details%%%%%%%%%%%%%%%%%%%%

\subsection{Experiment details}
\textbf{3DSeg-8 dataset}. 3DSeg-8 is an aggregate dataset from eight public medical datasets. It covers different organs/tissues of interest with either CT or MR scans, as shown in Table~\ref{tab1}. From each member dataset, we randomly select 90\% data to form the training set, and the rest 10\% as the test set. To improve the network robustness, we utilize three data augmentation techniques, including translation, rotation and scaling with the following settings: translating the data in any direction within 10\% margins of the target bounding box; rotating the data within [-5, 5]; and scaling the data within [0.8, 1.2] times of the original sizes.

\textbf{Network archietecture}. 
%We adopt the ResNet family (Resnet\_10, \_18, \_34, \_50, \_101, \_152, and \_200) architecture in 3D version as the backbone of Med3D networks. To avoid over downsampling, we set stride to 1 in blocks 3 and 4 with atrous convolutional layers as suggested in ~\cite{chen2018deeplab} to avoid signal decimation. For meet the requirement of segmentation, we replace the fully-connected layer with a 8-branch decoder where each branch consists of a $1 \times 1 \times 1$ convolutional kernel and a corresponding up-sample layer that scale the network output up to the original dimension.
We adopt the ResNet family (layers with 10, 18, 34, 50, 101, 152, and 200) and  pre-activation ResNet-200~\cite{he2016identity} architecture as the backbone of Med3D networks. To enable the network to train with 3D medical data, we modify the backbone network as follows: 1) we change the channel number of the first convolution layer from 3 to 1 due to the single channel volume input; 2) we replace all 2D convolution kernels with the 3D version; 3) we set the stride of the convolution kernels in blocks 3 and 4 equal to 1 to avoid down-sampling the feature maps, and we use dilated convolutional layers with rate $r=2$ as suggested in~\cite{chen2018deeplab} for the following layers for the same purpose; and 4) we replace the fully connected layer with a 8-branch decoder, where each branch consists of a 1x1x1 convolutional kernel and a corresponding up-sampling layer that scale the network output up to the original dimension.

\textbf{Training}. As each member dataset has a different number of training data, we take all the training data from the largest dataset and randomly augment the rest to generate an evenly distributed a balanced training dataset. We optimize network parameters using the cross-entropy loss with the standard SGD method~\cite{krizhevsky2012imagenet} where the learning rate is set to 0.1, momentum set to 0.9 and weight decay set to 0.001.

%%%%%%%%%%%%%%%%%%%%%%Data Magnitude on Med3D%%%%%%%%%%%%%%%%%%%%
\begin{figure}
\begin{center}
\includegraphics[width=0.46\textwidth]{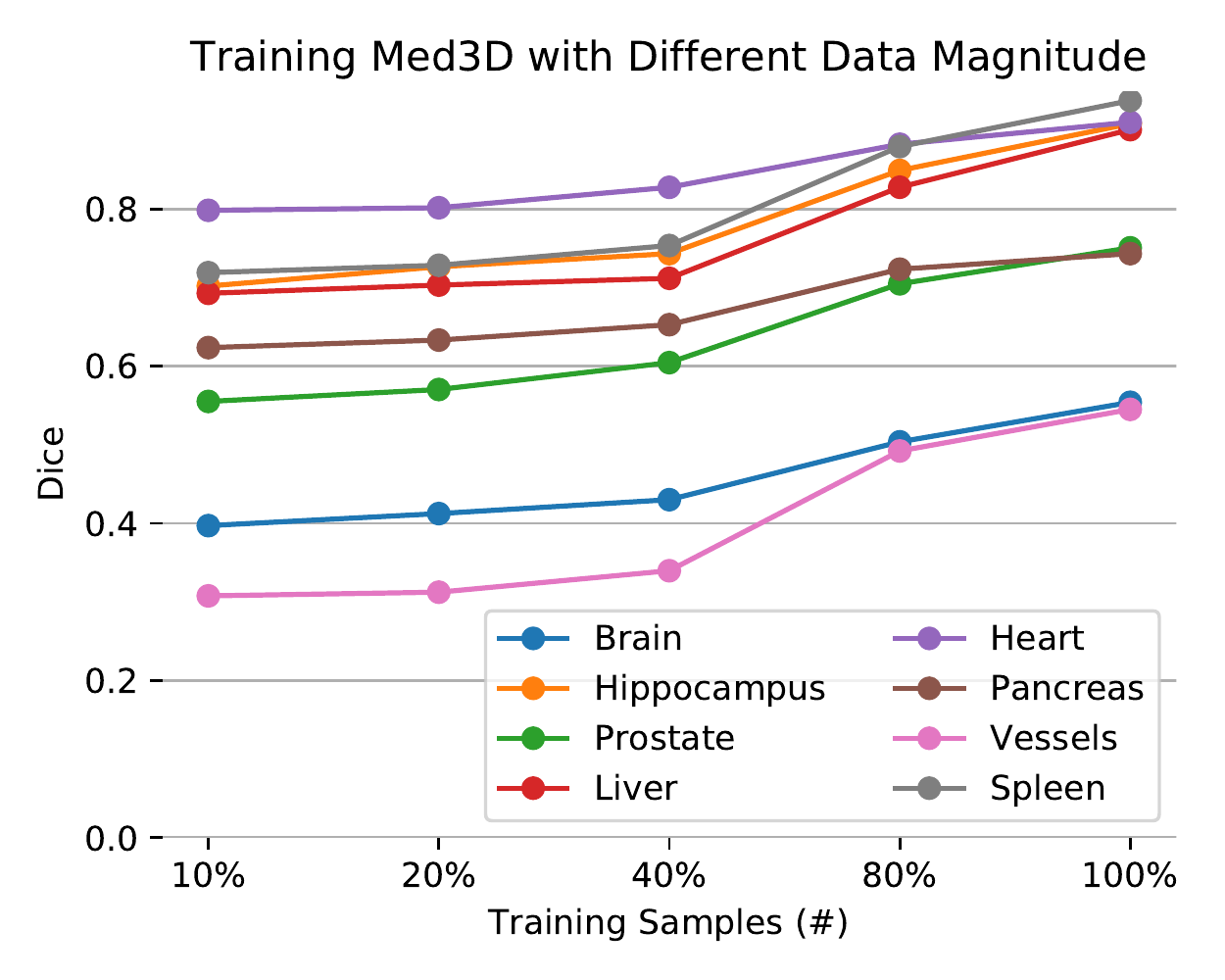}
\end{center}
   \caption{Random sample 10\%, 20\%, 40\%, 80\%, 100\% of training data and train Med3D.}
\label{fig4}
\end{figure}
%%%%%%%%%%%%%%%%%%%%%%Data Magnitude on Med3D%%%%%%%%%%%%%%%%%%%%

\textbf{Impact of training data magnitude}.
Sun et al.~\cite{sun2017revisiting} have found that the performance of vision systems increases logarithmically with the number of training data. To evaluate the impact of training data size in the medical imaging domain, we train a Med3D network (ResNet-152 backbone) with 10\%, 20\%, 40\%, 80\% and 100\% of the training data separately and use the same test set to compare the performance. As can be seen from Fig.~\ref{fig3}, the highest Dice scores for all tasks are obtained when the training data utilization is 100\%. When the amount of training data is only 10\% or 20\%, the model performance drops sharply due to the overfitting issue. When the number of training data grows, all the experiments show the same improvement trend, which indicates that the performance of Med3D and the magnitude of training data are also correlated.

%%%%%%%%%%%%%%%%%%%%%%Domains on Med3D%%%%%%%%%%%%%%%%%%%%
\begin{table}
\begin{center}
\caption{Train Med3D on single domain, two domains, four domains and eight domains.}
\begin{tabular}{c|c|c|c|c}
\hline
\multirow{2}{*}{Dataset} & \multicolumn{4}{c}{Dice} \\ \cline{2-5} 
 & One & Two & Four & Eight \\ \hline\hline
Brain~\cite{menze2015multimodal} & 54.67\% & 54.81\% & 54.83\% & \textbf{55.41\%} \\ \hline
Hippo~\cite{MSD} & 89.41\% & 89.45\% & 89.76\% & \textbf{90.86\%} \\ \hline
Prostate~\cite{MSD} & 72.58\% & 72.94\% & 73.29\% & \textbf{75.07\%} \\ \hline
Liver~\cite{lits} & 88.12\% & 88.75\% & 90.01\% & \textbf{90.11\%} \\ \hline
Heart~\cite{tobon2015benchmark} & 86.73\% & 88.16\% & 89.74\% & \textbf{91.05\%} \\ \hline
Pancreas~\cite{MSD} & 72.66\% & 73.81\% & 74.08\% & \textbf{74.33\%} \\ \hline
Vessel~\cite{MSD} & 53.98\% & 54.12\% & 54.49\% & \textbf{54.51\%} \\ \hline
Spleen~\cite{MSD} & 91.33\% & 92.08\% & 92.67\% & \textbf{93.80\%} \\ \hline
\end{tabular}
\label{tab4}
\end{center}
\end{table}

%\begin{figure}
%\begin{center}
%\includegraphics[width=0.43\textwidth]{latex/figures/domain.pdf}
%\end{center}
%   \caption{Train Med3D on single domain, two domains, four domains %and eight domains.}
%\label{fig5}
%\end{figure}
%%%%%%%%%%%%%%%%%%%%%%Domains on Med3D%%%%%%%%%%%%%%%%%%%%

\textbf{Impact of training set variety}.
We also set up a set of comparative experiments to investigate the Med3D performance with data from a various number of datasets. For example, we train Med3D with a solo member dataset and compare the results with the ones from two, four and eight member datasets. We use different colors to label the group settings for the experiments with two and four member datasets. As shown in Table~\ref{tab4}, Med3D reaches the highest performance when all the member datasets are used in training. This is because data varieties can provide complementary information and improve overall performance.

%%%%%%%%%%%%%%%%%%%%%%results comparison%%%%%%%%%%%%%%%%%%%%
\begin{table}
\begin{center}
\caption{Results of transfer Med3D to lung segmentation (Seg) and pulmonary nodule classification (Cls) with Dice and accuracy evaluation metrics, respectively.}
\begin{tabular}{c|c|c|c}\hline
Network & Pretrain & Seg & Cls \\ \hline\hline
\multirow{2}{*}{3D-ResNet10} & TFS & 71.30\% & 79.80\% \\ \cline{2-4} 
 & Med3D & \textbf{87.16\%} & \textbf{86.87\%} \\ \hline
\multirow{3}{*}{3D-ResNet18} & TFS & 75.22\% & 80.80\% \\ \cline{2-4} 
 & Kin~\cite{hara3dcnns} & 83.21\% & 82.83\%\\ \cline{2-4} 
 & Med3D & \textbf{87.26\%} & \textbf{88.89\%} \\ \hline
\multirow{3}{*}{3D-ResNet34} & TFS & 76.82\% & 83.84\% \\ \cline{2-4} 
 & Kin~\cite{hara3dcnns} & 85.82\% & 83.84\% \\ \cline{2-4} 
 & Med3D & \textbf{89.31\%} & \textbf{89.90\%} \\ \hline
\multirow{3}{*}{3D-ResNet50} & TFS & 71.75\% & 84.85\% \\ \cline{2-4} 
 & Kin~\cite{hara3dcnns} & 87.11\% & 74.75\% \\ \cline{2-4} 
 & Med3D & \textbf{93.31\%} & \textbf{89.90\%} \\ \hline
\multirow{3}{*}{3D-ResNet101} & TFS & 72.10\% & 81.82\% \\ \cline{2-4} 
 & Kin~\cite{hara3dcnns} & 88.32\% & 74.75\% \\ \cline{2-4} 
 & Med3D & \textbf{92.79\%} & \textbf{90.91\%} \\ \hline
\multirow{3}{*}{3D-ResNet152} & TFS & 73.29\% & 73.74\% \\ \cline{2-4} 
 & Kin~\cite{hara3dcnns} & 88.61\% & 75.76\% \\ \cline{2-4} 
 & Med3D & \textbf{92.33\%} & \textbf{90.91\%} \\ \hline
\multirow{2}{*}{3D-ResNet200} & TFS & 71.29\% & 76.77\% \\ \cline{2-4} 
 & Med3D & \textbf{92.06\%} & \textbf{90.91\%} \\ \hline
\multirow{2}{*}{3D-PreResNet200} & TFS & 70.66\% & 74.75\% \\ \cline{2-4} 
 & Med3D & \textbf{93.82\%} & \textbf{91.92\%} \\ \hline
\end{tabular}
\label{tab5}
\end{center}
\end{table}
%%%%%%%%%%%%%%%%%%%%%%results comparison%%%%%%%%%%%%%%%%%%%%
\subsection{Transfer learning experiments}
In the previous experiments, we have investigated the Med3D performance (ResNet-152  backbone) on the 3DSeg-8 dataset. The results demonstrate that the Med3D achieves the best segmentation performance when trained with the full 3D medical dataset that consists of eight domains. In this section, we explore the pre-trained Med3D network’s performance on an unseen dataset, to validate the transferable capability of the learned features. 

The 3DSeg-8 dataset consists of organs/tissues from eight different datasets that have moderate scale variations. To clearly reveal that the pre-trained network is generalized to be scale and task invariant, we conduct two experiments with data from lung regions, the lung segmentation Visceral dataset~\cite{jimenez2016cloud} and pulmonary nodule malignancy dataset LIDC~\cite{armato2004lung, armato2011lung}. We also compare results of using pre-trained Med3D with the ones of using pre-trained Kinetics model (Kin)~\cite{hara3dcnns}, and the ones of training from scratch (TFS). 

\textbf{Lung segmentation task}. We select the Visceral dataset because it includes abundant lung scans with four different imaging modalities/protocols which are unenhanced whole body CT, contrast enhanced abdomen and thorax CT, unenhanced whole body MR T1 and contrasted enhanced abdomen MR T1. 
There are 80 volumes in total and we pick 72 volumes for training and 8 volumes for testing. Both training and testing dataset contains 4 modalities. 

Lung segmentation task. We select Visceral dataset as it includes abundant lung segmentation annotations for the data in 4 different modalities. There are 80 volumes in total. We pick 72 volumes for training and 8 volumes for testing. Both training and testing data contains 4 modalities. During training, we use same segmentation architectures (ResNet family) for all comparison candidates. The major difference comes from the way how we initialize the segmentation network, using pre-trained Med3D, pre-trained Kin or TFS. When training with pre-trained models, we optimize the model parameters with Adam~\cite{kingma2014adam} starting from the 0.001 learning rate, while the learning rate of TFS is set to 0.01.
%%%%%%%%%%%%%%%%%%%%%%segmentation_training_loss%%%%%%%%%%%%%%%%%%%%
\begin{figure}
\begin{center}
\includegraphics[width=1.0\textwidth]{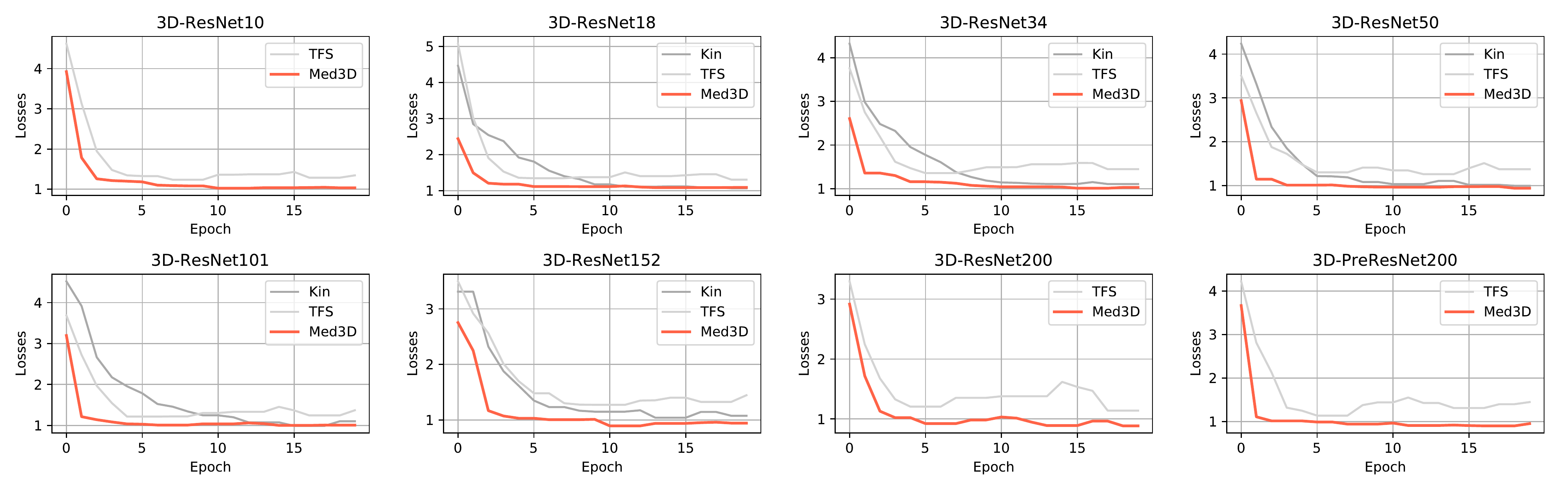}
\end{center}
   \caption{Training curve for lung segmentation.}
\label{fig5}
\end{figure}
%%%%%%%%%%%%%%%%%%%%%%segmentation_training_loss%%%%%%%%%%%%%%%%%%%%

As illustrated in Table~\ref{tab5}, all the results follow the same pattern that Med3D networks perform much better than Kin networks, and TFS networks are the worst. This confirms our initial assumption that there is a gap between 3D information captured by the temporal video data and the medical volume. To gain the best performance on 3D medical data, it is better to start with features capturing information about human physiological structures.

The sub-graphs in Fig.~\ref{fig5} demonstrate training curves of the segmentation task initializing differently. It can be seen that, with a certain amount of data and sufficient training iterations, all three networks converge to have stable losses. However, Med3D networks can push the loss even lower than the other two and show a much faster convergence speed.

%%%%%%%%%%%%%%%%%%%%%%classification_training_loss%%%%%%%%%%%%%%%%%%%%
\begin{figure}
\begin{center}
\includegraphics[width=1.0\textwidth]{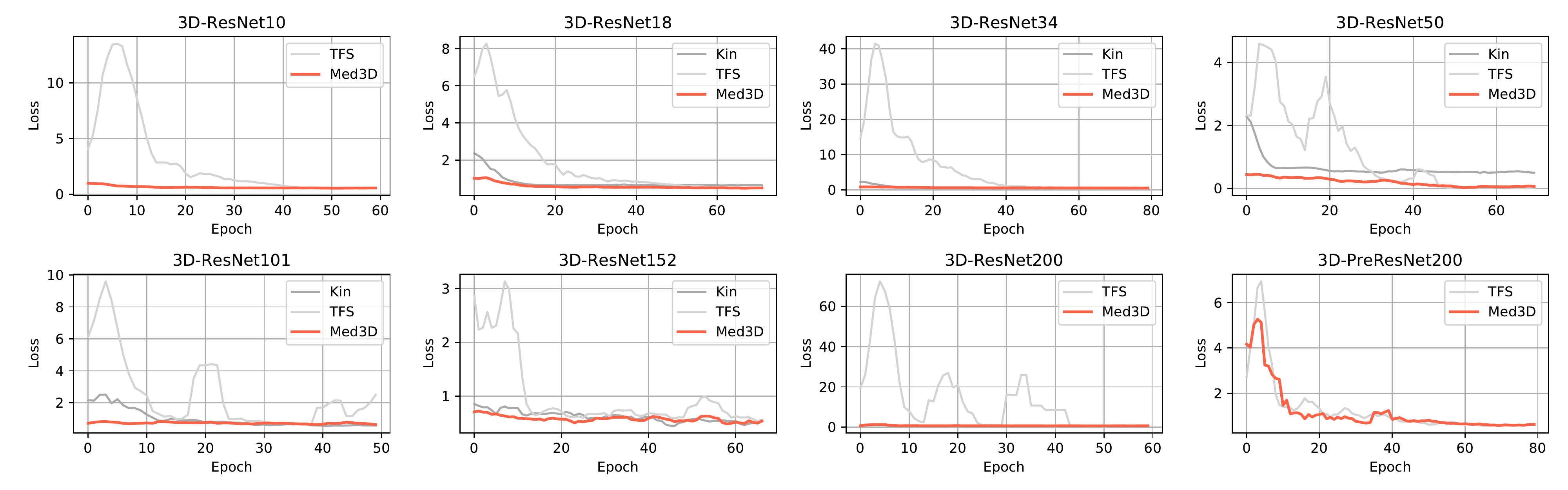}
\end{center}
   \caption{Training curve for pulmonary nodule classification. 
   %Where med, c3d and tfs represent pre-training model based on 3D medical dataset, pre-training model based on kinetics~\cite{kay2017kinetics}, and training from sctratch, respectively.
   }
\label{fig6}
\end{figure}
%%%%%%%%%%%%%%%%%%%%%%classification_training_loss%%%%%%%%%%%%%%%%%%%

\textbf{Pulmonary nodule classification task.} LIDC dataset~\cite{armato2004lung, armato2011lung} collected thoracic CT scans from 1,010 patients and nodules of each CT scan were annotated by four radiologists. The malignance of the nodules was defined in five levels, from benign to malignant according to the rules of LIDC-IDRI. We merge `1`, `2`, `3` as benign and `4`, `5` as malignancy for reducing subjective uncertainty suggested by ~\cite{han2015texture}. The goal of this task is to compare the network performance of using different initialization methods. We change the backbone network to a classification architecture by attaching full connected layers to the encoder. We use 1,050 nodules for training and 99 nodules for testing. The optimization parameters are consistent with the previous task.

As can be seen from Table~\ref{tab5}, results based on pre-trained Med3D networks significantly surpass the ones based on Kin and TSF. This demonstrates the effectiveness of the learned features of Med3D, which are also helpful for the classification task. Moreover, when the network depth is gradually increased, the performance of Med3D also increases. On the contrary, both Kin and TFS networks show declined performance when the network complexities are high. Together with the evidence shown in Figure 6 that the training losses of different networks are reduced to a similar level after long-enough training epochs, we can conclude that the extracted features from Med3D networks are better generalized for the classification task with a small set of data, while the other two methods show overfitting issues.  

Similar to Fig.~\ref{fig5}, Fig.~\ref{fig6} also provides the training loss curves that reflect the network convergence speed and grade of difficulty. As we can see from the figure, models trained from scratch have much higher fluctuation and converge much slower than the ones initialized with Med3D pre-trained weights. Since this task is a simple binary classification task, we can imaging that the network trained from scratch may have difficulties to converge for the complex classification tasks.

\subsection{LiTS challenge}
\label{litsdata}
In the previous section, we have demonstrated that the pre-trained Med3D networks have much better effectiveness and generalization for unseen data in both segmentation and classification tasks, compared to the networks pre-trained on natural scene videos and the ones trained from scratch. It makes us wonder if the Med3D can further boost the network performance on challenging tasks, such as LiTS challenge, where 3D networks trained from scratch usually have worse performance than 2D or 2.5D approaches using pre-trained models on natural images. 

The LiTS challenge dataset has totally 201 enhanced abdominal CT scans, which is further split to a training set with 131 scans and a test set with 70 scans. Only training data annotations are given to the public and the test ones are kept private with the host. The task is to segment the liver and liver tumors. It is known as a challenging task for two reasons. First, the number of training data is small. Second, the data is collected from different clinical sites with different scanners and protocols, causing large variations of the data quality, appearance and spacing. 

%The training data set contains 131 CT scans and the test data set 70 CT scans. 
%In order to segment liver, we replace the tumor label with liver label. In medical applications, data from different hospitals or different machines vary significantly in window width, window level, and pixel value distribution. This poses a huge challenge to the training and testing of the network. 
%We also set the window width range from -200 to 250 according to the CT Hounsfield unit value of the liver for data normalization.
%, and cut off the pixel value beyond the above range. In the range of the liver window, we remove the non-liver scan connected region around the liver, and in order to finally standardize the data range, we normalize the data into a normal distribution.

\begin{table}
\begin{center}
\caption{LiTS challenge results.}
\begin{tabular}{c|c|c|c}
\hline
Method & Dice & ASSD & Public \\ \hline\hline
TencentX & \textbf{96.6\%} & \textbf{1.0} & Unpublished \\ \hline
mastermind & 96.6\% & 1.0 & Unpublished \\ \hline
MILab & 96.0\% & 2.8 & Unpublished \\ \hline\hline
3D AH-Net~\cite{Liu20183D} & \textbf{96.3\%} & \textbf{1.1} & Published \\ \hline
H-DenseNet~\cite{Li2017H} & 96.1\% & 1.1 & Published \\ \hline
V-Net~\cite{milletari2016v} &93.9\% & 2.2 & Published \\ \hline\hline
Med3D & \textbf{94.6\%} & \textbf{1.9} & - \\ \hline
Kin &89.6\% & 5.0 & - \\ \hline
TFS &66.0\% & 30.4 & - \\ \hline
\end{tabular}
\label{tab6}
\end{center}
\end{table}

We set up the same segmentation network with ResNet-152 backbone and initialize it with the pre-trained Med3D. During training, we normalize all data with the average spacing value of the training data as some volumes do not give spacing information. We also normalize the intensity with a window width from -200 to 250 Hounsfield Unit. All the training hyper-parameters are same as the previous segmentation experiments. 

As shown in Table~\ref{tab6}, the Med3D network achieves 94.6\% Dice score without any post-processing, which is very close to state-of-the-art networks using ensemble techniques~\cite{Liu20183D, Li2017H}. We also compare the Dice results with pure 3D networks, such as V-net~\cite{milletari2016v}, Kin and TFS, and show that the Med3D outperforms those approaches with large margins. Similarly, the average symmetric surface distance (ASSD) also reflects that Med3D has superior segmentation results compared to other pure 3D approaches.  

\section{Conclusion}
In this work, we build a large-scale 3D medical dataset 3DSeg-8 and propose a novel framework to train Med3D networks with such data. The extracted features from Med3D networks are demonstrated to be effective and generalized, and can be used as the pre-trained features for other tasks with small training datasets. Compared with networks trained with natural videos or trained from scratch, the Med3D networks achieve superior results. We will release all pre-trained Med3D models as well as the related code. In the future, we will continue to collect more 3D medical data to further improve the Med3D pre-trained models.

\bibliographystyle{unsrt}  

\end{document}